# A TOPOLOGICAL METHOD FOR COMPARING DOCUMENT SEMANTICS

Yuqi Kong[1], Fanchao Meng[1] and Ben Carterette[2]

[1]Department of Computer & Information Sciences,
University of Delaware, Newark, USA
[2]Spotify, Greenwich Street, New York, USA

## ABSTRACT

*Comparing document semantics is one of the toughest tasks in both Natural Language Processing and Information Retrieval. To date, on one hand, the tools for this task are still rare. On the other hand, most relevant methods are devised from the statistic or the vector space model perspectives but nearly none from a topological perspective. In this paper, we hope to make a different sound. A novel algorithm based on topological persistence for comparing semantics similarity between two documents is proposed. Our experiments are conducted on a document dataset with human judges' results. A collection of state-of-the-art methods are selected for comparison. The experimental results show that our algorithm can produce highly human-consistent results, and also beats most state-of-the-art methods though ties with NLTK.*

## 1. INTRODUCTION

The problem focused in this paper is Document Semantics Comparison Problem as follows. Given two human readable, fairly long documents in similar lengths (and temporarily only in English), a real value to reflect the similarity between the two documents is desired. This problem is one of the fundamental problems lying at the heart of Natural Language Processing (NLP) and Information Retrieval (IR). To date, this problem has been attacked majorly from the statistical perspective, such as TF-IDF [1] [2] based methods, and vector space model (VSM) [3] methods. The former category directly utilizes TF-IDF information combined with statistical techniques to design the methods, while the latter category emphasizes and represents the relationships between words or constituents via VSM models, most of which are constructed still based on statistical information. In this paper, several state of the art and concrete such methods are selected for comparison. In Section 2, these methods will be briefly explained.

The methods proposed in this paper sheds a light from a different perspective, topology. More specifically, our methods are utilizing topological persistence [4] to represent the relationship between any two given documents, then the semantics similarity is computed from this representation. Our methods started with a very natural motivation. That is, if two documents have similar semantics, then they must have a relatively larger amount of relationships reflecting this similarity. Then a core task to formulate this similarity is to represent the relationships between the two documents. A document can be considered as a concrete carrier of its semantics. It consists of a collection of words and the relationships between these words. These relationships are ciphered in the grammars and the conventions in human languages. From a computational perspective, our arsenal to represent these relationships is actually quite limited. Then we need to be creative to take advantage of our weapons. Parse trees [5], constituency-based and dependency-based, are powerful tools to represent word relationships within a sentence. To measure the similarity or distance between any two words, a number of candidates are also ready to be picked. Then utilizing these two types of tools, we are able to partially construct the relationships in two documents. In other words, within a document, words in a sentence are organized in a connected component via a parse tree, and words in different document can be connected via checking the similarity or distance between them. In this way, a combinatorial graph is constructed, in which every path from word to another reflects a direct or indirect relationship, and the collection of such relationships can consequently reflect a portion of the relationship between the two documents. Then how to extract features from this graph becomes the key to define the similarity of semantics.

Topological persistence represents and extracts features of topological spaces from the algebraic topological space, and in algebraic topology language, it is also a one-dimensional abstract simplicial complex [4]. Our methods compute homology groups [4] and the corresponding topological persistence [6] for representing the relationships between the documents. The final similarity scores are computed based on the topological persistence. Again, in an intuitive way, the topological persistence for a given dimension contains a set of "holes" which have birth and death [6] so that their lifetime can be measured. The longer the lifetime, the more important the "hole". Moreover, in our case, a "hole" represents a piece of a sentence in one document has a relatively strong semantic relationship with a piece of another sentence in the other document. The background of topological persistence will be introduced in Section 3.

The major contribution of this paper is a novel algorithm to compute semantic similarity between documents. For the experimentation, we compare the similarity scores produced by our algorithm with those given by human judges, and also, we compare our algorithm with a collection of state-of-the-art methods. The experimental results strongly support that firstly our algorithm can produce similarity scores highly consistent with human judges; meanwhile it has better performance than most methods selected though ties with NLTK. Section 4 and 5 will present our algorithm and the experiments respectively. Discussion and conclusion follow in section 6 and 7.

## 2. RELATED WORK

The first method that we have particular interest is Doc2vec [7], which is a deep learning model designed base on Word2vec [8]. The essential idea of this method is actually not complicated. Since Word2vec is a model to represent words, then why not we add another vector (for paragraph) to represent document. Then the authors of Word2vec throw in another model named as Distributed Memory version of Paragraph Vector (DMPV) [8] which acts as a memory, in a rough way to explain, memorizing the topic of document. This model has demonstrated some significant progress on several NLP tasks such as topic extraction and sentiment analysis [8]. However, topic extraction and sentiment analysis are not equivalent tasks to document semantics comparison problem, since the latter problem is concentrating on the base level semantics (without any implications, metaphors, ironies or any other such concerns). Then it is also interesting to know if this state-of-the-art method would also work well on our problem.

The second work that has been selected is a state-of-the-art concrete software library for NLP research and development, named as NLTK [9]. In this library, a vector space model-based method [11] for comparing document semantics can be found. Its general idea is nothing more a classic one. that is, TF-IDF combined with cosine similarity, and the vector space model in NLTK is actually constructed via utilizing TF-IDF. However, it optimizes the procedure a lot via some statistic techniques, and in this way, its performance has been having a good reputation.

The third word is Text2vec [12]. This is one of the newest implementation libraries for NLP, and it contains four methods for comparing document semantics. They are Jaccard similarity [13], Cosine similarity (similar as the one provided by NLTK), Cosine with LSA [14] and Euclidean distance with LSA. These methods are all implemented based on vector space model and statistics. The most interesting point is that a couple of methods are combined with LSA which has been playing an important role in topic extraction.

We are not aware of any published work to date on use of topological persistence on representation of semantics and comparisons except a paper published in 2013 by Xiaojin Zhu [15]. The author shows a relatively preliminary application of persistent homology in natural language processing. The objective of this paper is use persistence to distinguish simple rhythmic literatures, and child or adolescent writings. Specifically, in the methods proposed in this paper, each paragraph is formulated as a bag-of-words vector, and the datasets are nursery rhymes, and child and adolescent writings; moreover, the use of topological persistence is limited to the number of "holes". Our methods to represent and compare semantics will be designed for more sophisticated and general cases.

## 3. THEORETICAL BACKGROUND

In this section, we provide a brief introduction to fundamentals of homology theory and persistence. Intuitively, the major task of homology theory is to describe "holes" in a geometric space from the algebraic perspective, and persistence describe how persistent these "holes" are.

We start with the formal definition of the most important concept for formulating geometric spaces from the algebraic perspective, namely abstract simplicial complex.

In an abstract simplicial complex, a $p$-simplex is a $p$-dimensional simplex (e.g. a line segment is a 1-simplex, or a triangle convex hull is a 2-simplex). A $p$-chain is a formal sum of a set of $p$-simplices, written as $\sum_k \alpha_k \sigma_k$, where $\alpha_k$ is a co-efficient in the ground field $\mathbb{F}$ and $\sigma_k$ is a simplex. The $p$-cycles and the $p$-boundary are all $p$-chains. They are defined by the boundary operator, denoted by $\partial_p$. The boundary operator maps $p$-chains to $(p-1)$-boundaries. For example, given a triangle-shape convex hull, the boundary operator takes this convex hull and returns the triangle consisting of three-line segments without the interior of the convex hull. The resulting triangle is called the boundary of the convex hull. A $p$-cycle is a $p$-chains to whom apply the boundary operator will return zero. In other words, $\partial_2 = 0$. This property of boundary operators is called the chain complex property [4] [16]. A $p$-homology-classes is a set of $p$-cycles equivalent to one another, and the equivalence relation is defined in the way that if two $p$-cycles $\mathcal{C}_p^i$ and $\mathcal{C}_p^j$ are equivalent then $\mathcal{C}_p^i - \mathcal{C}_p^j$ is a $p$-boundary. A $p$-homology-group is the set of $p$-homology-classes computed from a complex. In a formal way, the $p$-homology-group is defined as $\mathcal{H}^p = \frac{\mathcal{Z}_p}{\mathcal{B}_p}$, where $\mathcal{Z}_p$ is the $p$-cycle group, and $\mathcal{B}_p$ is the $p$-boundary group.

Equivalently, the homology *group* [4] is also defined as $\mathcal{H}^p = \frac{\mathcal{K}er(\partial_p)}{\mathcal{I}m(\partial_{p+1})}$.

A filtration is a sequence of indexed sets attached to the abstract simplicial complex, where each simplex is assigned with a filtration value [6] indicating the moment when this simplex is about to appear in the sequence, the birth of a $p$-homology-class is the earliest moment when any of its representative $p$-cycle appears in the sequence, the death of a $p$-homology-class is the earliest moment when a $p + 1$-cycle containing the exact set of vertices of any of this $p$-homology-class's $p$-cycle appears (N.B. this moment can be infinity), and the lifetime of a $p$-homology-class is the distance between its birth and death. A visualization of the collection of the set of $p$-homology-classes with birth-death pairs computed from an abstract simplicial complex with a filtration is called a persistence diagram.

## 4. ALGORITHMS

**Given:**

Two English documents, $D_i$, $D_j$.
A predetermined parameter, $\theta_t \in [0,1]$.
A stopwords list, $W_s$

**Seek:**

A real value reflecting the semantic similarity between $D_i$ and $D_j$.

**Step 1:** For each sentence, $S_{ik} \in D_i$, and each sentence, $S_{jh} \in D_j$, where $k, h$ are indices for the two sentences respectively, compute their dependency-based parse trees.

**Step 2:** Do tokenization and lemmatization on each parse tree, (i.e. prune non-word terminals and convert words to lemmas).

**Step 3:** For each term pair $(t_{ik}^p, t_{jh}^q)$, where $t_{ik}^p \in S_{ik}$, $t_{jh}^q \in S_{jh}$, and $t_{ik}^p, t_{jh}^q \notin W_s$, and $p, q$ are indices for the two terms respectively, compute the word similarity between $t_{ik}^p$ and $t_{jh}^q$ via utilizing Wordnet [10] LIN similarity [17]. We denote this similarity $\tau_t(t_{ik}^p, t_{jh}^q)$, and $\tau_t(t_{ik}^p, t_{jh}^q) \in [0,1]$.

**Step 4:** $\theta_w$ is taken as a threshold for the word similarities. For each $\tau_t(t_{ik}^p, t_{jh}^q)$, if $\tau_t(t_{ik}^p, t_{jh}^q) \geq \theta_t$, then the two terms $t_{ik}^p$ and $t_{jh}^q$ are considered as two vertices and placed into an empty graph, and also an edge between the two terms is created. The weight on this edge is $\tau_t(t_{ik}^p, t_{jh}^q)$.

**Step 5:** For each parse tree obtained in Step1, it is a graph, in which the vertices are the terms and the edges are determined by the tree. For each edge in this graph, set its weight to 1. Then union all such graphs obtained from the parse tree with all resulting graphs from Step 4. The final graph will be an undirected and weighted graph, denoted by $\mathcal{G}_{ij}$.

**Step 6:** The graph obtained from Step 5 is a one-dimensional abstract simplicial complex, denoted by $\Sigma_{ij}^1$. Given this abstract simplicial complex, compute the homology group for dimension one, denoted by $\mathcal{H}_{ij}^1$.

**Step 7:** Set the filtration value for each simplex in $\Sigma_{ij}^1$ via utilizing the weights in $\mathcal{G}_{ij}$ (i.e. the filtration value for an edge, $(t_{ik}^p, t_{jh}^q)$, which is a one-dimensional simplex is set to $1 - \tau_t(t_{ik}^p, t_{jh}^q)$, and the filtration values for the two corresponding vertices which are zero-dimensional simplices on this edge are all set to the same values.

**Step 8:** Given the homology group, $\mathcal{H}_{ij}^1$, and the abstract simplicial complex, $\Sigma_{ij}^1$, obtained from Step 6, and the filtration values for the simplices obtained from Step7, compute the topological persistence for dimension one, denoted by $\mathcal{P}_{ij}^1$.

**Step 9:** In $\mathcal{P}_{ij}^1$, for each homology class, denoted by $[c_l]$, its birth is determined by the minimum filtration value of the simplices in $c_l$; and its death is determined by the maximum filtration value which is equal 1. Compute the lifetime of $[c_l]$ which is equal to $\min_{(t_u^l, t_v^l) \in c_l} \{\tau_t(t_u^l, t_v^l)\}$, where $(t_u^l, t_v^l)$ is a one-dimensional simplex in $c_l$ (which is also an edge in $\Sigma_{ij}^1$).

**Step 10:** The final similarity between $D_i$, and $D_j$ is the sum of all lifetimes of the homology classes in $\mathcal{P}_{ij}^1$, (i.e. $\sum_{[c_l] \in \mathcal{P}_{ij}^1} \min_{(t_u^l, t_v^l) \in c_l} \{\tau_t(t_u^l, t_v^l)\}$), obtained from Step 9.

## 5. EXPERIMENTATION

**Design:** The goal of this experiment is evaluating the performance of Algorithm (TopoSem) proposed in Section 4. TopoSem will be compared to human judges. The performance of TopoSem should reflect how competent this algorithm can compare semantics of two documents as human judges. A dataset containing a collection of English documents will be utilized. For each pair of documents, human judges determine if this pair of documents have similar semantics, and provide a score to measure their similarity. Taking these similarity scores, two groups of document pairs can be constructed. One group contains all pairs that are determined by human judges as similar in semantics, and the other contains dissimilar pairs. Then TopoSem is applied to both groups to give each document pair in the two groups a semantic similarity score. If these scores given by TopoSem agree on the two groups, then the performance of TopoSem is considered as positive. Essentially, this experiment is a classification task, where the two classes are determined by the two groups, and TopoSem will be tested on classifying document pairs collected from the two groups into the two classes.

**Settings:** The dataset in use is provided by Michael D. Lee [18] which contains 50 documents selected from the Australian Broadcasting Corporation's news mail service. The lengths of documents vary from 51 to 126 words, and cover a number of broad topics. The documents in this dataset have been evaluated by human judges. For each pair of documents, there is an average of scores from human judges ranging from 1 to 5, where 1 indicates highly unrelated, and 5 indicates highly related. Besides this 50-document dataset, Michael D. Lee also provides an additional dataset containing 300 background documents which are in average longer than the 50 documents.

To utilize WordNet [10] Lin [17] to compare word meanings, an information content database needs to be specified. What is selected in this experiment is SemCor provided by WordNet 3.0.

Group 1, denoted by $\mathbb{G}_1$, of document pairs is constructed by collection all pairs that are scored $\leq$ 2.5, and Group 2, denoted by $\mathbb{G}_2$, is constructed by collecting all pairs scored $\geq$ 3.5. $\mathbb{G}_1$ contains 1095 pairs, and $\mathbb{G}_2$ contains 46 pairs. Since the rest of pairs could be hardly determined as similar or dissimilar even by human judges, then they are not considered in our experiment. The predetermined parameter $\theta_t$ takes values 1, 0.95, 0.9, 0.85 and 0.8 for five trials. The stopwords list in use is provided by Onix Text Retrieval Toolkit [19] which contains 571 words.

**Method:** TopoSem is applied to both $\mathbb{G}_1$ and $\mathbb{G}_2$ to compute a similarity score for each document pair. For each group, the 95% confidential interval of the scores is computed, denoted by $\mathcal{I}_c^1$ and $\mathcal{I}_c^2$ respectively, where the superscripts are indices. $\mathbb{G}_1$ is set as the positive class (setting $\mathbb{G}_2$ as the positive class is also tested). If a document pair in $\mathbb{G}_1$, denoted by $d_1(x, y) \in \mathbb{G}_1$, where 1 is the group index, x, y are document indices, is given a score, denoted by $\alpha(d_1(x, y))$, which holds $\alpha(d_1(x, y)) \leq \mathcal{U}(\mathcal{I}_c^1)$, then $d_1(x, y)$ is considered as a true positive, where $\mathcal{U}(\cdot)$ takes the supremum of a given interval. If $\alpha(d_1(x, y)) \geq \mathcal{L}(\mathcal{I}_c^2)$, then $d_1(x, y)$ is a false negative, where $\mathcal{L}(\cdot)$ takes the infimum of a given interval. Similarly, for a document pair $d_2(a, b) \in \mathbb{G}_2$, if $\alpha(d_2(a, b)) \geq \mathcal{L}(\mathcal{I}_c^2)$, then $d_2(a, b)$ is considered as a true negative; and if $\alpha(d_2(a, b)) \leq \mathcal{U}(\mathcal{I}_c^1)$, then $d_2(a, b)$ is considered as a false positive. Foreach trial (with a specific $\theta_t$), all true positives, true negatives, false positives and false negatives are collected and counted, and then precision, recall and F1 score are calculated.

Table 1. Error rates for TopoSem and control groups of methods

| Methods | $\mathbb{G}_1$ Error Rate | $\mathbb{G}_2$ Error Rate | Average Error Rate |
|---|---|---|---|
| TopoSem ($\theta_t = 1.00$) | **2.19 %** | 19.56 % | 10.88 % |
| TopoSem ($\theta_t = 0.95$) | 2.37 % | 19.56 % | 10.97 % |
| TopoSem ($\theta_t = 0.90$) | 3.01 % | 19.56 % | 11.28 % |
| TopoSem ($\theta_t = 0.85$) | 4.29 % | 17.39 % | 10.84 % |
| TopoSem ($\theta_t = 0.80$) | 6.48 % | 21.73 % | 14.11 % |
| Doc2Vec | 33.33 % | 17.39 % | 25.36 % |
| Text2vec (Jaccard | 15.07 % | 39.13 % | 27.10 % |
| Text2vec (Cosine | 9.50 % | 39.13 % | 24.32 % |
| Text2vec (Cosine + | 9.50 % | 32.61 % | 21.01 % |
| Text2vec (Euclidean + | 7.03 % | 32.61 % | 19.82 % |
| NLTK | 2.28 % | **15.22%** | **8.75 %** |

Table 2. Precisions, Recalls and F1 scores with Group 1 as positive and Group 2 as negative.

| $\mathbb{G}_1$ as positive and $\mathbb{G}_2$ as negative | | | |
|---|---|---|---|
| Methods | Precision | Recall | F1 |
| TopoSem ($\theta_t = 1.00$) | 0.97 | 0.99 | **0.98** |
| TopoSem ($\theta_t = 0.95$) | 0.97 | 0.99 | **0.98** |
| TopoSem ($\theta_t = 0.90$) | 0.96 | 0.99 | 0.97 |
| TopoSem ($\theta_t = 0.85$) | 0.94 | 0.99 | 0.96 |
| TopoSem ($\theta_t = 0.80$) | 0.91 | 0.99 | 0.95 |
| Doc2Vec | 0.60 | 0.99 | 0.75 |
| Text2vec (Jaccard | 0.84 | 0.98 | 0.91 |
| Text2vec (Cosine | 0.89 | 0.98 | 0.94 |
| Text2vec (Cosine + LSA) | 0.89 | 0.98 | 0.93 |
| Text2vec (Euclidean + | 0.91 | 0.98 | 0.94 |
| NLTK | 0.96 | 0.99 | **0.98** |

Table 3. Precisions, Recalls and F1 scores with Group 2 as positive and Group 1 as negative.

| $\mathbb{G}_2$ as positive and $\mathbb{G}_1$ as negative | | | |
|---|---|---|---|
| **Methods** | **Precision** | **Recall** | **F1** |
| TopoSem ($\theta_t = 1.00$) | 0.59 | 0.35 | 0.44 |
| TopoSem ($\theta_t = 0.95$) | 0.59 | 0.33 | 0.43 |
| TopoSem ($\theta_t = 0.90$) | 0.63 | 0.31 | 0.42 |
| TopoSem ($\theta_t = 0.85$) | 0.62 | 0.22 | 0.32 |
| TopoSem ($\theta_t = 0.80$) | 0.50 | 0.12 | 0.20 |
| Doc2Vec | 0.55 | 0.03 | 0.05 |
| Text2vec (Jaccard | 0.25 | 0.04 | 0.06 |
| Text2vec (Cosine similarity) | 0.31 | 0.07 | 0.12 |
| Text2vec (Cosine + LSA) | 0.44 | 0.10 | 0.17 |
| Text2vec (Euclidean + LSA) | 0.50 | 0.16 | 0.25 |
| NLTK | 0.70 | 0.39 | **0.50** |

In this experiment, a control group of methods are also tested on the same task. The methods include Doc2vec, Text2vec and NLTK. One of the implementations of Doc2vec (whose name is Gensim [20]) provides a direct interface to compare semantics of two documents. Text2vec provides Jaccard similarity, cosine similarity [22], cosine similarity with TF-IDF, cosine similarity with LSA and Euclidean distance with LSA these methods for comparing document semantics directly. NLTK provides a vector space model based on TF-IDF, then the document similarity can be computed via cosine similarity.

**Experimental Results:** The experimental results are listed in Table 1, Table 2 and Table 3. Table 1 shows the error rate for each method. In this table, the winner for $\mathbb{G}_1$ is our TopoSem with $\theta_t = 1.00$ while NLTK and our TopoSem with $\theta_t = 0.95, 0.90$ produce similar results. The winner for $\mathbb{G}_2$ is NLTK while our TopoSem with $\theta_t = 0.85, 1.00, 0.95, 0.90$ also produce similar results. In average, the winner is NLTK and our TopoSem with $\theta_t = 0.85$ is the second winner. The difference between the average error rate of NLTK and that of TopoSem with $\theta_t = 0.85$ is 2.09% which is not significant. It can be observed that the error rates on $\mathbb{G}_2$ are higher the error rates on $\mathbb{G}_1$. The reason is that the dataset is skewed so that the misclassified pairs in $\mathbb{G}_2$ impact the error rates on $\mathbb{G}_2$ much more significantly than the misclassified pairs in $\mathbb{G}_1$.

Table 2 shows the precision, recall and F1 score for each method under the case that $\mathbb{G}_1$ is set as positive and $\mathbb{G}_2$ is set as negative. The winner of F1 score is our TopoSem with $\theta_t = 1.00, 0.95$ and NLTK. Table 3 shows the case that $\mathbb{G}_1$ is set as negative and $\mathbb{G}_2$ is set as positive. In the latter case, NLTK is slightly better than our TopoSem but still not significant. Additionally, the reason that the F1 scores in Table 3 are lower than those in Table 2 is again because of the skew in the dataset.

## 6. DISCUSSION

From table 1, deep learning methods such as Doc2vec and Text2vec have much worse error rate compare with our method and NLTK. Since those two deep learning methods require massive training documents to pretrain the model. If in the scenario that lacks such pretraining dataset, such as Michael D. Lee's dataset we used which only contains 300 background documents. The performance of those methods will hurt. Contrarily, non-training methods such as our topological method and NLTK are capable in any scenario.

Can our method do better? This was the first question we asked ourselves right after the results popped out. Since the performance of our method seems does not significantly better then NLTK and even slightly worse in the case that $\mathbb{G}_1$ is set as negative and $\mathbb{G}_2$ is set as positive. The bottleneck comes from Wordnet and LIN. They are far out of date tools, the number of synsets in Wordnet is not adequate, those out-of-vocabulary words compromised the performance by hindering the formation of simplicial complex (aka, meaningful "holes"). Furthermore, LIN may not be the best option either. WordNet provided LIN as its out-of-box word similarity algorithm.

However, our major goal is to propose a unique novel topological structure that can unify both syntactic and lexical semantics of the document and quantify the semantics without any pretraining procedure. The results proved the validity of our proposal. Moreover, there is a lot of room for improvement.

## 7. CONCLUSION

In this paper, a novel algorithm for comparing document semantics is proposed. This algorithm is designed based on topological persistence, which is distinguished from most methods for the same task. The experimental results provide strong support to our algorithm showing that it can unify both syntactic and lexical semantics of documents, then produce highly human-consistent results, and also outperform some state-of-the-art methods.

## 8. FUTURE WORK

Although, TopoSem shows potentials that it is highly consistent with human judgment. The results indicated the performance does not significantly outperform the control methods. There are many aspects that we can do to improve this novel approach. A new version of the algorithm is under development. We are plan to involve parse tree trimming to trim unnecessary nodes in order to reduce the effect of noise homology classes. For the current algorithm, we only use filtration value to weight terms edges formed in step 4. In the next version of the algorithm, we try to not only weigh the terms edges but also parse tree edges then use harmonic mean to combine two types of weights together. We hope this could give TopoSem a more comprehensive similarity function. Furthermore, one of the limitations of TopoSem is complexity, this impedes the application of TopoSem to large datasets. After the algorithm matured, we will focus on the optimization of TopoSem.